%% file: main.tex
\documentclass[10pt]{article} 
\usepackage[preprint]{tmlr}

\input{math_commands.tex}

\usepackage{graphicx}%
\usepackage{multirow}%
\usepackage{amsmath,amssymb,amsfonts}%
\usepackage{amsthm}%
\usepackage{mathrsfs}%
\usepackage[title]{appendix}%
\usepackage{xcolor}%
\usepackage{textcomp}%
\usepackage{manyfoot}%
\usepackage{booktabs}%
\usepackage{algorithm}%
\usepackage{algorithmicx}%
\usepackage{algpseudocode}%
\usepackage{listings}%
\usepackage{hyperref}

\usepackage[most]{tcolorbox}

\usepackage{soul}

\usepackage{tikz}
\usetikzlibrary{positioning, shapes.geometric}

 \newcommand{\indep}{\perp\!\!\!\perp}
 \newcommand{\nindep}{\perp\!\!\!\perp\!\!\!\!\!\!\!/ \,\,\,}

\title{A Causal Framework for Mitigating Data Shifts \\ in Healthcare}

\author{\name Kurt Butler$^{1,2}$ \email kbutler2@ed.ac.uk \\
    \name Stephanie Riley$^{1,3}$ \email stephanie.riley-3@manchester.ac.uk \\
    \name Damian Machlanski$^{1,2}$ \email d.machlanski@ed.ac.uk \\
    \name Edward Moroshko$^{2}$ \email emoroshk@ed.ac.uk \\
    \name Panagiotis Dimitrakopoulos$^{1,2}$ \email pdimitra@ed.ac.uk \\
    \name Thomas Melistas$^{2,4,5}$ \email melistas.th@gmail.com \\
    \name Akchunya Chanchal$^{1,6}$ \email akchunya.chanchal@kcl.ac.uk \\
    \name Konstantinos Vilouras$^{2}$ \email konstantinos.vilouras@ed.ac.uk \\
    \name Zhihua Liu$^{1,2}$ \email zliu7@ed.ac.uk \\
    \name Steven McDonagh$^{1,2}$ \email s.mcdonagh@ed.ac.uk \\
    \name Hana Chockler$^{1,6}$ \email hana.chockler@kcl.ac.uk \\
    \name Ben Glocker$^{1,7}$ \email b.glocker@imperial.ac.uk \\
    \name Niccol{\`o} Tempini$^{1,8}$ \email N.Tempini@exeter.ac.uk \\
    \name Matthew Sperrin$^{1,3}$ \email matthew.sperrin@manchester.ac.uk \\
    \name Sotirios A. Tsaftaris$^{1,2}$ \email S.Tsaftaris@ed.ac.uk \\
    \name Ricardo Silva$^{1,9}$ \email ricardo.silva@ucl.ac.uk\AND
    \addr $^{1}$Causality in Healthcare AI Hub (CHAI), UK \\
    $^{2}$The University of Edinburgh, UK; $^{3}$University of Manchester, UK \\
    $^{4}$Athena Research Center, Greece; $^{5}$NKU Athens, Greece \\
    $^{6}$King's College London, UK; $^{7}$Imperial College London, UK \\
    $^{8}$University of Exeter, UK; $^{9}$University College London, UK
}

\def\month{MM}  
\def\year{YYYY} 
\def\openreview{\url{https://openreview.net/forum?id=XXXX}} 

\begin{document}

\maketitle

\begin{abstract}
Developing predictive models that perform reliably across diverse patient populations and heterogeneous environments is a core aim of medical research. However, generalization is only possible if the learned model is robust to statistical differences between data used for training and data seen at the time and place of deployment. 
Domain generalization methods provide strategies to address data shifts, but each method comes with its own set of assumptions and trade-offs. To apply these methods in healthcare, we must understand how domain shifts arise, what assumptions we prefer to make, and what our design constraints are. 
This article proposes a causal framework for the design of predictive models to improve generalization. Causality provides a powerful language to characterize and understand diverse domain shifts, regardless of data modality. This allows us to pinpoint why models fail to generalize, leading to more principled strategies to prepare for and adapt to shifts. 
We recommend general mitigation strategies, discussing trade-offs and highlighting existing work. Our causality-based perspective offers a critical foundation for developing robust, interpretable, and clinically relevant AI solutions in healthcare, paving the way for reliable real-world deployment.
\end{abstract}

\section{Main}\label{sec1} 
Consider a hypothetical group of researchers that deploy a sepsis prediction model. At the hospital where it was developed, the model has impressive accuracy, passes various diagnostic tests and performs well during cross-validation. At a different hospital, later investigators found that the model was no longer accurate, and its predictions differed significantly from what doctors expected. 
It was observed that the distributions of the input data  differed between institutions. Standard corrections like importance weighting are attempted to resolve the issue, but performance does not significantly improve. What went wrong? Later investigators notice that the two hospitals use different monitoring equipment, meaning that the same patient will produce different data depending on the device used. When the measurement process changes, the relationship between vital signs and sepsis is not preserved across sites. The correct solution was measurement calibration, not importance weighting. 
This example highlights a general problem that statistical descriptions of data shift can be ambiguous about their causal origin. In such cases, we must ``map out" the system to identify solutions, and a purely statistical map will not do. To describe the processes that generate data, and to describe how those processes are affected by interventions, measurement recalibrations, and other exogenous influences, we require a causal map.

Although our example is fictional, Machine Learning (ML) models are increasingly being widely deployed in real life.
Specific examples include the QRISK model for the incidence of cardiovascular disease \citep{HippisleyCox2017_QRISK}, the PREDICT model for breast cancer survival \citep{Wishart2010_PREDICT}, and the automated retinal disease
assessment (ARDA) for diabetic retinopathy \citep{brant2025performance}. 
There is already a huge number of new healthcare and clinical prediction models that are proposed every year \citep{arshi2025number}. Instead of contributing to the ever-growing avalanche of new models, we advocate for the development of models that are robust, reusable and that can be shared across institutions and applied in different settings \citep{van2025enemies}.
However, variations in how data are generated affect the statistical relationships that are used in prediction.
Variations in disease prevalence across population strata can lead to misdiagnoses \citep{leeflang2009diagnostic, manrai2016genetic}. Variations in equipment, staffing, or protocols across time may result in noisy, missing, or otherwise mismatched data \citep{VanCalster2023}. Generally, variations may occur due to regional differences \citep{hwang2026landscape} and the passage of time \citep{vela2022temporal}. Whether these variations matter is context dependent, and it may be unclear a priori as to what types of variations a model will be sensitive to.
Models may be naturally robust to certain variations, but in other cases ignorance to such effects will lead to suboptimal prediction in new settings.

These variations, which we call data shifts, are variously referred to as data set shifts \citep{subbaswamy2018counterfactual}, domain shifts \citep{zhang2021adaptive}, or distribution shifts \citep{fang2020rethinking}, across medicine, computer science and statistics, with occasional minor differences in meaning.
It can be challenging to navigate this fragmented research landscape, and deciding how one should move from problem identification to technical solutions remains unclear. Every problem is unique in that the prediction target, the variables available to be used in prediction making, and the foreseeable data shifts may all vary across problems. Even in clinical research, existing reporting guidelines like TRIPOD+AI request that investigators report how predictors are defined and chosen, but the underlying logic as to why those predictors are chosen is left open-ended for the investigators \citep{collins2024tripod+}. Establishing a clear, structured framework for this transition is essential to ensuring that the resulting predictive models are fair, transparent, and interpretable for doctors, engineers, administrators and the public.

In this paper, we propose causality as a unifying framework to address data shift problems and to guide the design of robust models. Inspired by recent works in causality \citep{jalaldoust2024partial, jones2024causal, castro2020causality}, health research \citep{Subbaswamy2020,subasri2025detecting, Hickey2013, sperrin2022targeted} and ML \citep{salaudeen2025domain, richens2024robust}, we argue that causality is an essential tool for the design of predictors that can generalize well. Our contributions are as follows:
\begin{enumerate}
   \item  We propose a conceptual framework for the design of predictive models intended for uses that involve data shifts. We advocate for a case-by-case approach to designing predictors, acknowledging that problem-specific conditions and requirements often lead to design preferences.
   \item We explore how causal knowledge can allow us to determine different classes of problems that require different solutions, and we explore one version of this approach using augmented causal graphs. 
   \item We identify key pieces of information that one should extract from a problem description to determine the problem class that will shed light on applicable solutions. 
\end{enumerate}
Although healthcare and clinical research serve as rich sources of inspiration for our framework,  our ideas are broadly applicable to data shift problems in other disciplines.

\section{Sources of Variation in Healthcare and Medicine}
\label{sec:healthcare}
Before discussing our framework, it is useful to have an intuition for how data shifts arise. For the purposes of this work, we consider data sets that describe populations of patients interacting with healthcare systems, potentially over a period of time. From a healthcare perspective, the ways in which moving from one context to another can affect data are innumerable. Even within the same geographical region, health institutions may differ radically in staffing, instrumentation and in how they collect data \citep{subasri2025detecting}. Data shifts also can occur due to the passage of time, as seasonal changes, epidemics, policy changes, and general changes in societal behaviour can also affect the observed data sets. Some sources of data shifts may be straightforward to locate, such as the onset of the COVID-19 pandemic, but other shifts may have unobservable, unknown or undocumented origins.

To facilitate thinking about data shifts from the health perspective, we propose three broad categories. Our goal in this section is to group shifts by their source of variation. This differs from other categorizations, which often group shifts by their mathematical properties \citep{zhou2022domain}. These categories are not intended to be exhaustive or mutually exclusive. Rather, they serve as a heuristic to guide the process of identifying sources of variation in health domains, which is an important part of our problem. 

\paragraph{Population-level shifts}
This category covers fundamental changes in the underlying population of patients. These include demographic shifts, which encompass changes in the distribution of age, sex, ethnicity, socioeconomic status, deprivation, or other factors. These also include case-mix shifts, where disease prevalence, severity, duration or the presence of comorbidities may vary. This category can also account for referral and access biases, where strata-dependent access to health institutions affects which patients are likely to appear in data sets. 

Examples of population-level shifts are common in the literature. 
Demographic changes between training and deployment can exacerbate existing health inequities by leading to less accurate diagnoses or estimates of risk for people who already struggle to access the necessary care \citep{Ganapathi2022, jones2024causal}. Furthermore, a prediction model developed using hospital admission data may not generalise well to primary care. Patients admitted to hospital are typically more unwell or may have already undergone prior clinical filtering, such as a general practitioner (GP) referral. A model developed in hospital settings may therefore be miscalibrated when applied in primary care, where patients tend to be earlier in the disease course \citep{Knottnerus2002}. 

\paragraph{Exposure, selection and observation-level shifts}
Apart from changes in the patient population itself, some shifts occur in the processes by which institutions interact with, treat, and record data from the population. This category concerns the process of recording data and its contingencies on the resources, practices, protocols and staffing of the hospitals in which data is collected.

Observation-level shifts concern changes in the way in which data is collected. For instance, a model may be trained on objectively measured Body Mass Index (BMI) data recorded in a clinical setting, yet is deployed in a setting where only less-accurate, self-reported BMI is available \citep{gosse2014accurate}. Subjective and score-based tests are also vulnerable. The Glasgow coma scale (GCS) is a subjective measure of consciousness, and GCS scores for the same patient can vary depending on who performs the observation  \citep{mclernon2014glasgow}. In instrument-based analyses like radiology or spectroscopy, differences in hardware calibration can produce unique artifacts and biases that do not generalize \citep{bratchenko2025overestimation, glocker2019machine}. Even software updates can periodically induce data shifts that affect the performance of computerized medical imaging systems \citep{roschewitz2023automatic}. Ignoring shifts of this type risks mistaking artifacts of the observation process as spurious indicators of change in patients' health.

Exposure-level shifts, in contrast, reflect variations in how different hospitals prioritize treatment and admission.
In clinical research, differences in cohort study design can lead to differences in observed outcomes \citep{pye2018assumptions}. A classic example from oncology is the ``Will Rogers'' phenomenon, where the introduction of more-sensitive diagnostic tools appears to trigger a shift in the data, where patients move between disease stages without any actual change in their health or treatment \citep{feinstein1985will}. 

Finally, in most healthcare and clinical research there is a risk of \textit{selection bias}, where data shifts occur because we implicitly only record data from a subset of the population. Selection biases go under various names in practice, such as healthy user bias (where individuals with proactive health-seeking behaviors are systematically overrepresented) or access bias (where systemic or strata-dependent barriers influence which patients appear in the data). The issue of selection bias is a general and important one, and the connections to causality have been explored in recent work \citep{mathur2025simple}.

\paragraph{Definition-level shifts}
We distinguish a third category to account for shifts in the semantics of health.  While population shifts concern the subjects and observation/exposure shifts concern the measurement process, definition-level shifts involve fundamental changes in how clinical states are quantified, codified, and understood. Definition-level shifts are not due to a change in data quality, but due to a more fundamental shift in what the data means. Such shifts can happen when updates to institutional procedures, national guidelines or international standards occur, potentially rendering statistical relationships in historical data obsolete \citep{sahiner2023data, cohen2024subtle}. Shifts of this level may also occur across linguistic divides, particularly for language models. For example, models trained on Western medical data fail to adapt to local dialects or cultural descriptions of symptoms in low and middle income countries \citep{ong2026large}.

In the simplest case, clinical definitions may periodically change or evolve. The redefinition of clinical criteria for sepsis, from Sepsis-2 to Sepsis-3, fundamentally altered the labeled population by moving from inflammatory to organ-dysfunction markers \citep{singer2016third}. In the case of emerging diseases, existing definitions might not be sufficient to record them. COVID-19 did not have an assigned ICD-10 diagnosis code until April 2020, with earlier presentations of the disease being recorded as pneumonia or acute respiratory distress syndrome \citep{kadri2020uptake}. 
Without accounting for these shifts, models risk becoming ``time-locked'' to outdated clinical standards.

\section{Understanding How Shifts Arise in Data}
Data shifts arise because the real life processes which generate data are affected by environment-specific conditions or by external disruptions. As we discussed earlier, an environment might refer to a particular hospital, a particular time, a particular study, or potentially even an intersection of these. Sometimes the data used for training is chosen out of convenience or accessibility, rather than being perfectly matched to where the predictive model is eventually deployed \citep{sperrin2022targeted}. The problem is not a lack of data either, because even models trained on large datasets can be susceptible to data shifts as long as the deployment environment differs from the training environment. To make sense of this, it helps to use a graphical notation to describe how data shifts interact with systems of variables. We employ the augmented causal graph (ACG) notation, which we provide a primer for in Box 1.

\subsection{A primer on augmented causal graph notation}

Causal modeling allows us to answer questions about how data was generated. 
Compared to associational models, which tend to represent relationships between \textbf{variables} in terms of probabilities, causal models make explicit the directional dependencies between variables according to the order of events that generate their values. The traditional way to represent such ordered dependencies is using causal graphs \citep{pearl2009causality}. For prediction problems, we may write out nodes corresponding to each variable of interest, represented below by circles. Sometimes we may be aware of (or speculate the existence of) \textbf{latent variables} which we cannot directly observe. We include latent variables in the graph, perhaps denoting them with an unshaded circle or a dashed outline. Finally, for the purposes of prediction problems, one of the observed variables should be designated as the \textbf{prediction target}, colored orange. 

\begin{center}
\includegraphics[width=6cm]{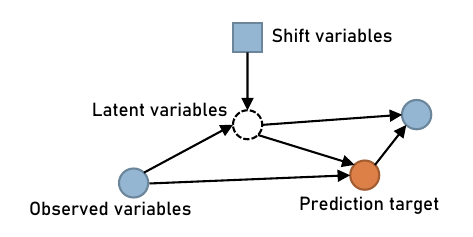}
\end{center}

Edges in the causal graph express direct dependencies between variables. That is, if there is an edge $A \to B$ between two variables $A$ and $B$, then this means that forcing $A$ to different values should affect the likelihood of $B$, with all other factors held constant. Importantly, the \textit{absence} of edges in the graph are used to express \textit{conditional independence}, where $B$ is independent of its non-descendants conditioned on its parents. For example, if $B \leftarrow A \to C$, then $B$ is independent of $C$ when conditioned on their common cause $A$, denoted $B \indep C |A$. Conversely, causal graphs also inform us of \textit{collider bias}, where conditioning on a common descendant of two variables induces a correlation, $X \nindep Y | Z$, that does not exist normally, $X \indep Y$, \citep{pearl2009causality}.

Using circles and edges as above, we can describe how variables connect and influence each other in a causal manner. However, a data shift may interrupt such a causal system, removing some dependencies or modifying others. We adopt an extension of the standard causal graphs \citep{dawid2021}. Denoted using a square, \textbf{shift variables} are used to index various environments (also called settings, domains, regimes, or contexts in other literature). An environment is a state of the data-generating process, which we assign a number to without placing a probability distribution over said numbers. For example, it could be useful to annotate datasets that were recorded pre- and post-COVID-19 with different labels, such as 1 or 2, but it is not generally useful to model a probability distribution for these labels.  Edges from a shift variable to another variable express a sensitivity to the change of setting. If there is an edge $S \to C$, where $S$ is a shift variable, then this expresses that the probability distribution $p_S(C)$ can vary as a function of the value of $S$. To make this more concrete, consider the following graph.

\begin{center}
\includegraphics[width=6cm]{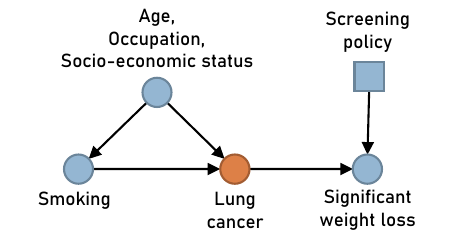}
\end{center}

In the graph above, we are interested in predicting the diagnosis of lung cancer. It is well-established that smoking causes increased risk of lung cancer, but the exact nature of that relationship is confounded by age, occupational hazards, and socio-economic status. Patients of lung cancer also typically experience significant weight loss. 
At present, people aged 55 -- 74 who have ever smoked are screened for lung cancer. Otherwise, patients are only tested for lung cancer if they present to their GP with symptoms. Consider the screening criteria was widened to include both a wider range of ages and people who have never smoked. This could then cause some patients to be diagnosed sooner and at early stages of the disease. As such patients may not be as sick and the average rate of weight loss changes. In this case we may see a data shift in the weight loss variable. In graphical notation, this means there is a single edge from the screening policy to significant weight loss.

\subsection{Classical domain generalization}
\label{sec:domain_generalization}
In machine learning, much of the work on handling data shifts falls under the literature on Domain Generalization (DG). Works in this area follow the language of supervised learning, framing the data shift problem in terms of covariates $X$ (inputs to the predictive model) and target $Y$ (the output variable). From this formulation, four basic shift types emerge, each corresponding to changes in different components of the joint distribution $P(X,Y)$ \citep{scholkopf2012causal}. It is common in this literature to describe probability distributions as being either \textit{unstable} (significantly altered by the shift) or \textit{stable} (also called invariant, meaning the effect of the shift is negligible).

Under covariate shift, the distribution of inputs $P(X)$ changes while the conditional $P(Y|X)$ remains stable. For example,~a model trained on younger patients encountering an older population at deployment. Concept shift describes the opposite situation where $P(X)$ stays constant but $P(Y|X)$ evolves, e.g., 
revised treatment protocols can change the mapping from patient features to outcomes.
The alternative factorization yields two additional types. Firstly, label shift occurs when outcome prevalence $P(Y)$ changes but how outcomes manifest in features $P(X|Y)$ remains stable, e.g., when disease rates differ across hospitals but symptoms present identically. Secondly, anti-causal features shift is the converse: $P(Y)$ is preserved but $P(X|Y)$ varies, e.g., when different imaging equipment produces different representations of the same underlying pathology.

\begin{figure}
    \centering
    \includegraphics[width=14cm]{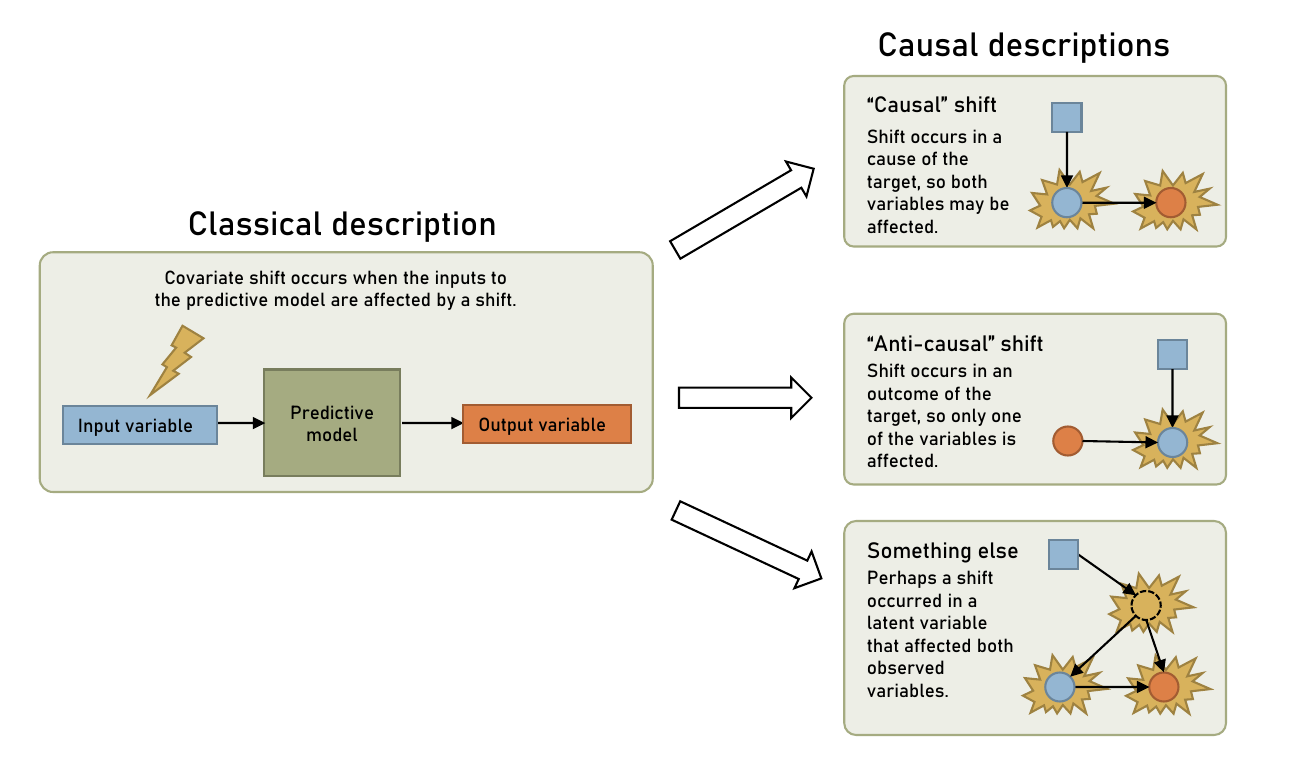}
    \caption{An illustration of how classical domain generalization differs from the causal theory of domain generalization. In this example, we show that even in a simple setting, a \textit{covariate shift} could arise from several distinct causal scenarios.}
    \label{fig:covariate-shift}
\end{figure}

This taxonomy, while useful, is agnostic to causal structure. It groups all inputs as ``covariates" without distinguishing whether $X$ causes $Y$, is caused by $Y$, or relates to $Y$ only through confounding. We visualize this distinction in Fig. \ref{fig:covariate-shift}. A shift in $P(Y|X)$ could arise because the causal mechanism changed, because a latent confounder distribution shifted, or because selection bias distorts the observed relationship, each producing the same statistical signature but requiring different solutions. Causal reasoning resolves this ambiguity by modeling the data-generating process itself, allowing us to anticipate which relationships will remain stable and diagnose the source of observed shifts.

Even in the classical view, it is not necessary that every domain shift problem falls neatly into one of these four types.
Joint shifts represent the most complex and realistic category of distribution shifts, where multiple components of $P(X,Y)$ change simultaneously. Unlike the ``pure'' shift types described above, joint shifts involve combinations of shifts that can create intricate patterns of model degradation that are difficult to diagnose and address with single-strategy approaches. For example, a shift can involve both a change in $P(X)$ and in $P(Y|X)$, creating compound effects that cannot be addressed by adaptation strategies designed for either covariate shift or concept shift alone. Indeed, joint shifts are more common than not in healthcare, making real-world deployment uniquely challenging and favoring model design around stable predictors.

Confounder shifts represent a particularly important subtype of joint shifts~\citep{Reddy2025} in healthcare. These occur when latent variables $U$ that influence both $X$ and $Y$ change the distribution. Even when the causal mechanisms $P(X|U)$ and $P(Y|X,U)$ remain constant, a change in $P(U)$ propagates through the causal graph to simultaneously affect $P(X)$, $P(Y)$ and $P(Y|X)$.

To address shifts in practice, there are many classes of methods within the DG literature. \textit{Test-time} or \textit{online adaptation} methods allow the systems to perform automated adjustments to self-correct during deployment \citep{wang2021tent, wang2022continual}. When retraining models during deployment is possible, models can be designed to actively adapt to the new environment as new data is received.  Domain \textit{adaptation} is a similar concept where unlabeled deployment-environment data is used to improve the original training, but the model is not updated during later deployment \citep{gong2012geodesic, kulis2011you,ganin2015unsupervised}. 
These approaches are not always feasible due to privacy regulations such as the European General Data Protection Regulation (GDPR) \citep{tian2023privacy}, since a retrained model might memorize personal information \citep{hartley2023neural}. Other challenges are that retraining consumes power, memory and time that might not be available during deployment. 
\textit{Invariant representation learning} methods attempt to generalize by mining for patterns that generalize across multiple environments, either using multiple datasets or expert knowledge \citep{arjovsky2020invariantriskminimization, muandet2013domain}. 
\textit{Meta learning} methods sit between these extremes, aiming to design models that will be quick to adapt to new environments by leveraging several environments seen during training \citep{hospedales2021meta}.  

\subsection{What causality can tell us about shifts}
\label{sec:causality_for_dg}
Where shifts enter the causal graph matters. If a shift affects how a symptom is produced by a disease, then that symptom cannot be used to detect the presence of said disease. In contrast, if the shift occurs only in the prevalence of this disease, then predictions about the symptoms can still be made using the old model, up to some loss of statistical efficiency. In causal language, these two prediction tasks are different. One is prediction in the ``causal'' direction (predicting an outcome from its causes) and one is prediction in the ``anti-causal'' direction (predicting an outcome from its symptoms or downstream effects).\footnote{Note this can only be done for diagnostic, not prognostic, models.} 

\begin{figure}
    \centering
    \includegraphics[width=0.9\linewidth]{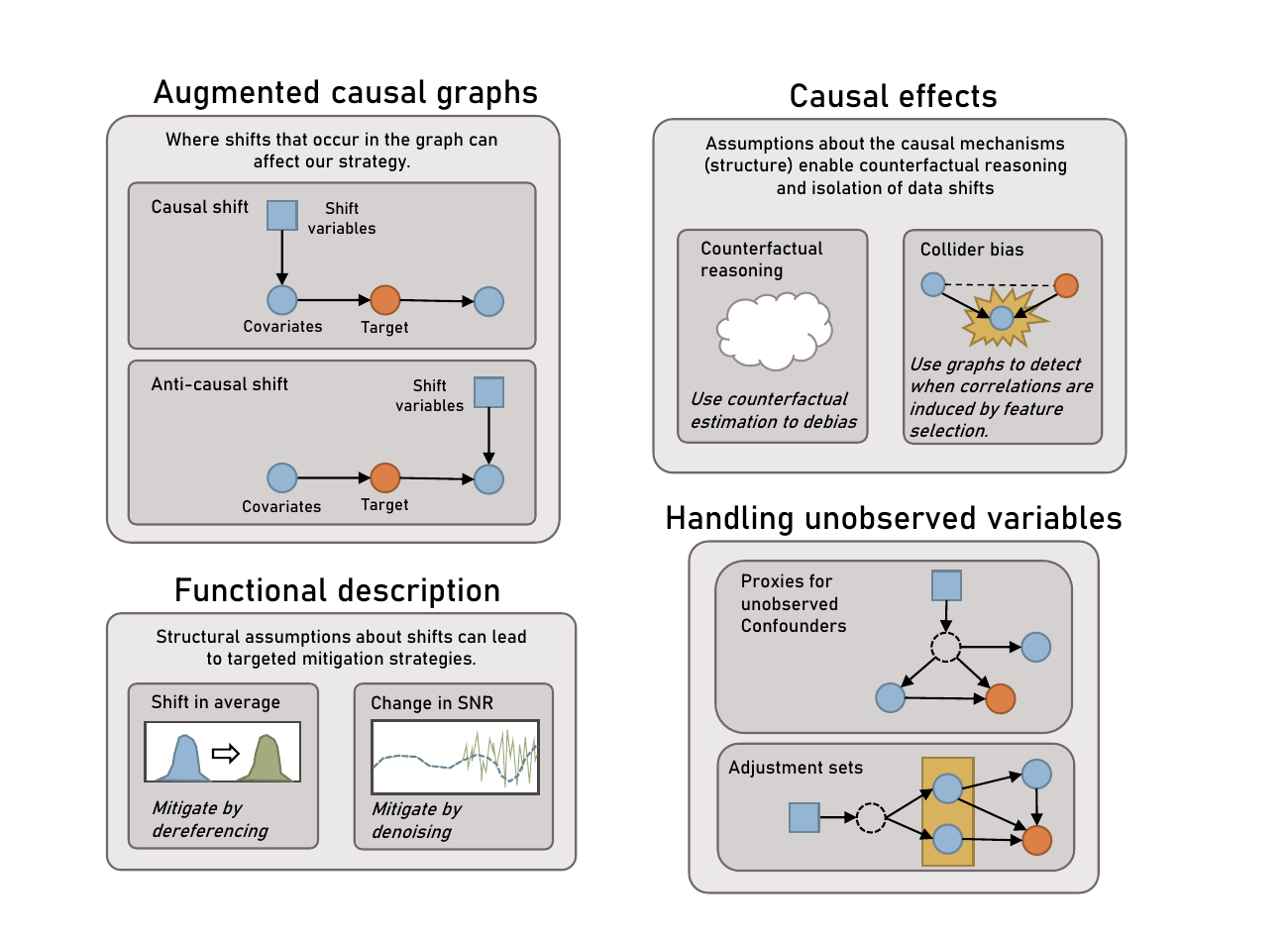}
    \caption{An illustration of various ways in which causal descriptions enhance our understanding of data shifts and inform our strategies to address them. Causal descriptions, either as graphs or descriptions as a data-generation process, are useful to mitigate data shifts. As we later show in Fig. \ref{fig:roadmap}, formulating causal descriptions is a crucial part of our framework.}
    \label{fig:glossary}
\end{figure}

In our graphical notation, these correspond to two separate cases, shown in Fig. \ref{fig:glossary}. 
The emphasis here is not that shifts fall neatly into distinct categories. In real world systems, there might be multiple shifts, or a single shift might affect both causes and effects of a disease. These cases are more complex than the basic ones outlined above. However, all of these examples underline the fact that causality matters for DG, and causal reasoning helps us consider how these shifts ultimately affect a predictive model. If we expect a model to generalize under a sufficiently large set of shifts, then causal reasoning becomes a necessity for robust generalization \citep{richens2024robust}. 

A principal advantage of causal reasoning in this setting is that it permits us to be proactive in our model design~\citep{subbaswamy2018counterfactual}. This allows us to reason about the effects of data shifts before we observe them. But it comes at the price of specifying the details about the anticipated shifts, which will likely come from domain knowledge. This is not necessarily a downside as this strategy gives domain experts an opportunity to leverage their knowledge to build bespoke models tailored specifically to the data they have. The exact form in which knowledge should be encoded to help with building models can vary, but we illustrate several examples in Fig. \ref{fig:glossary}.

In the causality literature, several methods for the data shift problem have been proposed~\citep{Pearl2011,zhang2015multisource,peters2016causal,magliacane2018domain,subbaswamy2018counterfactual,arjovsky2020invariantriskminimization,Kaur2022,Subbaswamy2022,wang2023causal,tsai2024proxy,jalaldoust2024partial}. The method of \textit{invariant causal prediction} attempts to achieve generalization by learning causal mechanisms that are unaffected by data shifts \citep{peters2016causal}, barring interventions that affect the target itself. 
\textit{Data augmentation} methods are used when certain features might be useful for prediction, but are known to be not universally applicable (e.g., detecting polar bears by looking for snowy images). Such methods are especially useful in computer vision, where tools such as image segmentation and inpainting give clear strategies for data augmentation \citep{ouyang2022causality, duan2024causal}. \textit{Counterfactual data augmentation} is a sub-approach that specifically uses a causal model to formulate how to edit images \citep{melistas2024benchmarking}. 
Other methods take a strongly analytic approach, analyzing the transportability of causal relationships across environments using detailed graphical models \citep{jalaldoust2024partial}. While useful when the model is right, we do not take the stance that this is the only way to proceed. Causal graphs in real systems are often misspecified or incomplete, and data shifts need not behave exactly as our simple mathematical descriptions. However, as we observe in the case of "causal" and "anti-causal" shifts, even basic graphical information can be useful to inform strategies for generalization.

\subsection{Other forms of causal information}
The graphical descriptions afforded to us by ACGs allow us to visualize where shifts occur and what their effects might be, but additional information, when available, can greatly benefit solving the problem. 
For example, the strength of a data shift often matters. If a data shift is not very strong, then many distinct deployment strategies may work well, because ultimately there was not much adaptation required from the model. This property has been demonstrated empirically and theoretically in simple models \citep{salaudeen2025domain}. 
Also, the mathematical or functional form of the shift is useful to know. For example in electroencephalography, a form of data shift frequently occurs due to a wandering reference voltage \citep{yao2019reference}. However, because the data shift is assumed to take on the form of a localized additive bias, which affects all observations within nearby brain regions similarly, techniques such as bipolar or common average montages are used during preprocessing to negate the effects of the data shift \citep{acharya2019overview}. 

In general, there is no universal algorithm that can address data shifts, but when extra information is available, it can lead to specialized solutions for data shifts as well. Because every problem may differ in the form that shifts take, the data that is available, and other environment-specific idiosyncrasies, the design of robust predictive models in practice is an open-ended pursuit.

\section{Mapping Problems to Solutions}
\begin{figure}
    \centering
    \includegraphics[width=0.95\linewidth]{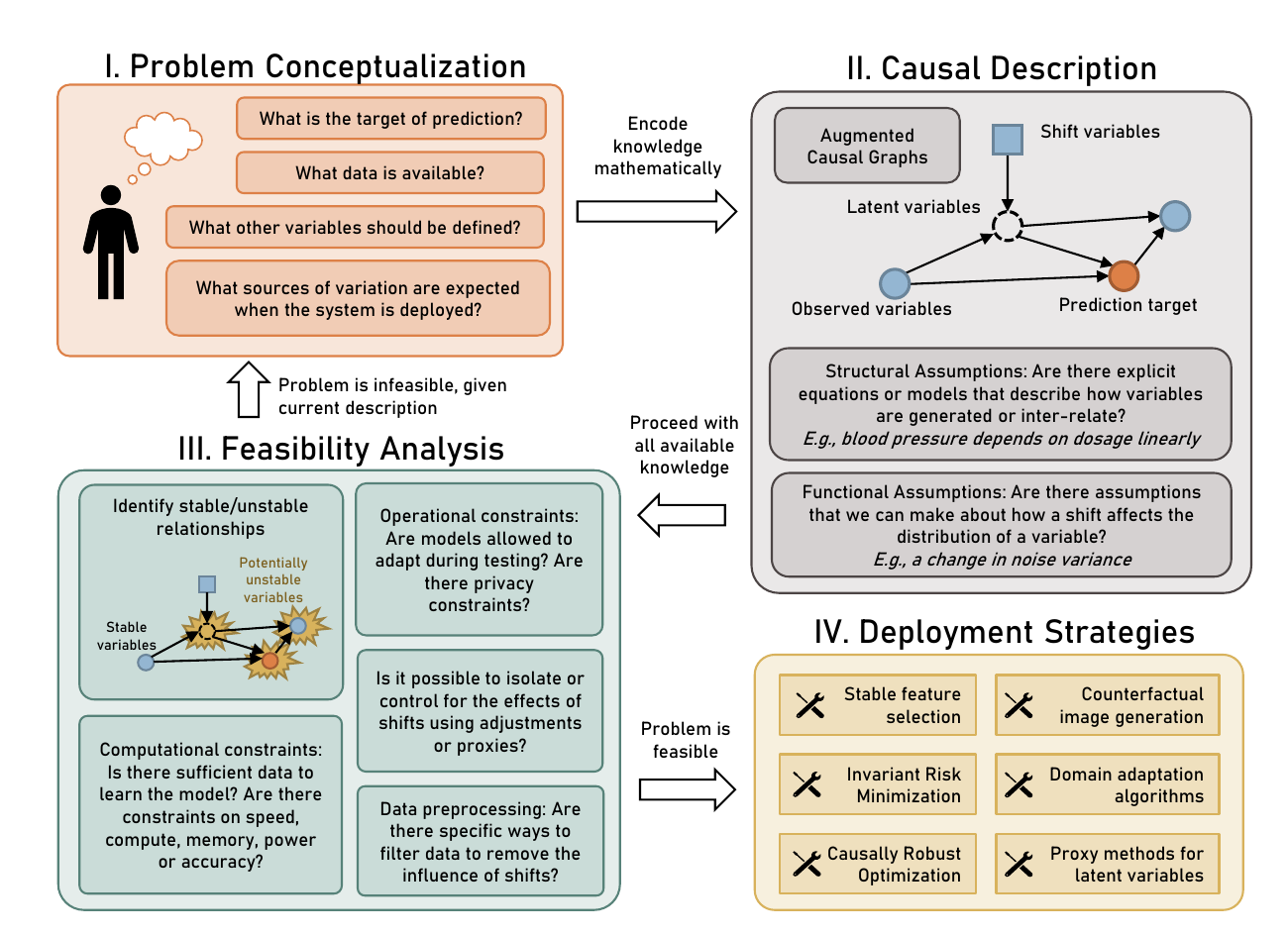}
    \caption{
    A roadmap of our framework for developing predictive models. We define four stages: I.~Problem conceptualization, II.~Causal description, III.~Feasibility analysis, and IV.~Deployment strategies. While the stages can be followed sequentially, model design is iterative and may require revisiting earlier stages as we evaluate the feasibility of the model.}
    \label{fig:roadmap}
\end{figure}

Given a prediction problem, our goal is to clarify the process of thinking carefully and arriving at a solution. By reasoning causally throughout this process, we can better anticipate how the system will behave when deployed, and ultimately when it is exposed to data shifts. Depending on the situation, there might be many adequate solutions, or there may be none. In the case that there are many solutions, preferences might only emerge after further contemplation, and these preferences could also be context dependent. Our goal is not to provide a formal algorithm to select a given solution for dealing with data shifts, but to produce a general framework for how these solutions are proposed, evaluated for feasibility, and compared. We visualize the roadmap for our framework in Fig. \ref{fig:roadmap}. We break down the process into four stages: Problem Conceptualization, Causal Description, Feasibility checks, and Deployment Strategies.

\paragraph{I. Problem Conceptualization}
Our focus so far has been on predictive modeling with machine learning models, but often in healthcare and medicine, predictive modeling is a means to an end. Predictive models may be used to inform decision making, anticipate negative events, and evaluate multiple possible interventions, but the exact way that these models are designed and used is flexible. Some goals, such as forecasting patient trajectories, can be naturally formulated as predictive tasks. Other problems may be more indirectly connected to predictive tasks, and so there is flexibility in how the predictive model may be used in problem solving. The goal then is to think through how prediction is used, and what the likely sources of variation are that may affect the results, with both clinical and engineering expertise in the conversation. 

During this stage, one must pick their prediction target, and then specify what inputs will be used to make predictions from available data repository. Is any important variable missing that we should collect? Additionally, one must consider what environments the predictive model is intended to be making predictions in. 

\paragraph{II. Causal Description}
After specifying a predictive task, the next step is to describe the problem mathematically. The coarsest description that one can give is to split variables into covariates and targets. As discussed in previous sections and Fig. \ref{fig:covariate-shift}, there are several reasons that more detail in the problem formulation, specifically causal details, can improve the deployment of predictive models.

Depending on the problem, more or fewer assumptions about the model may be appropriate. Sometimes, we might only have access to a (hypothesized) augmented causal graph, but even this basic information is useful, as shown in Fig. \ref{fig:glossary}. When available, additional information can be of further use. Assuming specific equations, functional forms or probability distributions are useful to improve inference, when the assumptions are appropriate. Information about the nature of a shift, or the strength of a shift, even if unknown, can be useful to devise strategies to mitigate their negative effects. 

\paragraph{III. Feasibility Analysis}
Given a causal description of the system, one can then check their model for feasibility. In general, there are many criteria that could be used to evaluate a model's utility, even at a high level. Constraints on the speed, model type, computer memory, compute, number of parameters, accuracy, false positive or negative rate, power consumption, interpretability, user privacy or data availability all matter in determining whether a given predictive model can be used.  If the required infrastructure, support, or resources needed for a long-term or sophisticated solution are not available, then the preferences for solutions will differ from institutions that have a larger budget and more investment in the system. Deciding on the best solution is an iterative process, requiring a revision of the modeling assumptions and desiderata across multiple rounds. 

As a first step, we need to check that there are enough observed variables to make predictions. Variables for which we only have noisy or proxy observations may not be usable for prediction. Also, it is often useful to identify the existence of stable variables for prediction, which are those unaffected by data shifts. Empirical studies in benchmarking predictive models suggest that unstable relationships are not necessarily useless for prediction \citep{salaudeenaggregation}, but their predictive utility cannot be verified a priori. Again, this relates to the idea that a data shift may also bias a prediction. Sometimes this bias is not strong enough to abandon a given predictive model for a different one that might need to make stronger assumptions about the nature of possible shifts. Still, stable relationships are useful to generate a baseline for predictive modeling, providing a lower bound on performance during deployment. 

Additionally, there are two categories of solutions that we can distinguish, adaptive and non-adaptive. For adaptive solutions, the AI model is permitted to use data after deployment to adapt itself to the new environment. Broadly, this includes paradigms like adaptive filtering \citep{rupp2016advances} and continual learning \citep{liu2020learning}, and the procedure of adaptation could involve changing the entire model's parameters (as seen in methods such as Elastic Weight Consolidation \citep{kirkpatrick2017overcoming}) or merely adapting a few specific parameters of the system. Non-adaptive solutions are those that do not receive updates while deployed. Not adapting to a new user can be beneficial to avoid costs in speed and compute related to processing new data, and also pose no privacy risk for new users.

\paragraph{IV. Deployment Strategies}
Once the desiderata for the problem are known, and details about our modeling assumptions and available data are given, one can invoke various methods to design a system that deploys well. As discussed in earlier, there are many strategies for doing this. Methods based on feature selection \citep{li2022invariant} or data augmentation \citep{volpi2018generalizing, von2021self, gao2023out} can be used to exclude unstable features from the model, or to adjust for imbalances in the training data, respectively. When multiple environments are available in the training data, the principle of Invariant Risk Minimization seeks to learn a single predictor for all observed training environments \citep{arjovsky2020invariantriskminimization}. Alternatively, one may leverage causal and statistical knowledge to attempt to optimally transport from one environment to a specific target environment \citep{jalaldoust2024partial, blanchet2025distributionally, sagawa2019distributionally}. The ultimate preference depends both on the feasibility requirements identified in Stage III and the causal assumptions and details expressed in Stage II.

In practice, multiple candidate solutions may be of interest. In such case, approaches of ensembles or cross validation are useful to produce a single prediction from a set of many. Additionally, during validation, multiple candidate solutions may be compared and evaluated empirically, producing information about their actual performance and costs that are not available in a purely theoretical analysis.

\subsection{Limitations of this Framework}
A core assumption of this framework is that data shifts admit mathematical descriptions using causal graphs and other mathematical structures. In domains like epidemiology, the process of creating graphs is entirely routine to some researchers \citep{hossain2025utilizing}. In other cases \citep{ong2026large}, it might be unclear how shifts can be monitored, detected or even hypothesized.

For shifts in data distributions, annotation practices, or technical infrastructure that can be formally characterized, causal AI methods offer rigorous frameworks for modeling and mitigation. While causal graphs provide valuable tools for understanding certain types of shifts, the performance of AI systems in healthcare emerges from complex socio-technical systems that extend beyond what can be captured in formal models. Shifts occurring at organizational (clinical protocols, workflow configurations), institutional (policy targets, resource allocation mechanisms), and socio-cultural levels (professional norms, epistemic values) could be equally consequential for system performance. These factors operate through mechanisms that might resist formal mathematical representation. Recognizing these as legitimate sources of shift does not diminish the superior value of causal modeling. 

This suggests that causal AI modeling, requiring input from experts and the use of local knowledge and domain knowledge, might need to work hand in hand with an analysis of the institutional arrangements, professional capabilities, and organizational processes necessary for well-aligned local deployment. What cannot be formally modeled might still be systematically addressed if sizable shifts that are difficult to characterize are present. This is the concern of other work we are carrying out in parallel \citep{tempini2025aishift} and is not a compromise, but a recognition that system performance in heterogeneous healthcare contexts depends on factors operating across multiple scales and modalities.

\section{Case Studies}
In this section, we consider three instances of our framework being applied to problems that appear in the healthcare literature.

\subsection{Case study on skin lesion classification with computer vision}

Consider a predictive model for melanoma detection from skin lesion images. Deep learning models achieve remarkable accuracy in this task \citep{esteva2017dermatologist}, but often fail to generalize across clinical sites due to differences in protocols, e.g., marking practices \citep{Winkler2019-vc, chamarthi2024mitigating}. This degradation commonly arises when models rely on spurious correlations between image artifacts (such as rulers or clinical markings) and disease labels, rather than learning clinically meaningful features of the lesion itself \citep{bissoto2019constructing, bissoto2020debiasing}. While these correlations can be highly predictive in the training dataset, they might not hold in other data distributions. Below, we show how our causal framework proactively addresses such data shifts.

\begin{enumerate}
\small
    \item \textbf{(Problem Conceptualization) }
    We observe images of skin lesions $X$, from which we make predictions for the malignancy of the lesion, provided by the biopsy result $Y$. The underlying disease state $U$, which is unobserved, manifests through the skin appearance $S$.
    During examination, clinicians may add markings $M$ to highlight areas that seem suspicious or warrant closer attention, which become part of the captured image. To model variation across clinical environments, we introduce an environment variable $D$ representing the institution or the treating physician, reflecting differences in marking conventions (some dermatologists may mark suspicious areas more moderately than others) and acquisition practices. Models are trained on data from one or more sites that follow specific practices, and must generalize to new sites where these conventions may differ.
    \item \textbf{(Causal Description)} 
    Now, we must map out all the causal pathways in our system. 
    The disease state $U$ determines skin appearance $S$, which is captured in the image: $U \to S \to X$. Clinicians observe the skin appearance $S$ and, based on this observation, decide which areas to mark: $S \to M$. However, marking conventions also depend on the doctor's individual practice patterns or institutional protocols: $D \to M$. These markings then appear in the final image: $M \to X$. As illustrated below, $D$ may also be correlated with $Y$ due to differences in referral or biopsy practices, creating a shortcut that models can exploit.
    A model may therefore learn to associate markings with malignancy rather than relying on lesion appearance, leading to performance degradation when marking conventions change \citep{castro2020causality}.

\begin{center}
        \includegraphics[width=6cm]{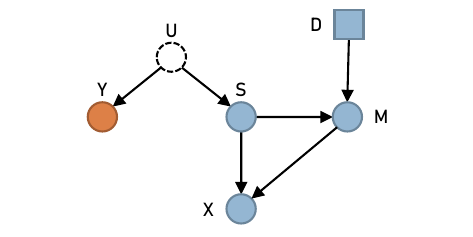}
\end{center}

    \item \textbf{(Feasibility Analysis)}
    The graph distinguishes stable from unstable relationships. The pathway $Y \leftarrow U \to S \to X$ represents the stable signal we aim to learn. In contrast, the pathway $D \to M \to X$ is unstable because marking strategies vary across institutions. A robust model requires mitigating the influence of $M$ on $X$, either by detecting and removing artifacts during data preprocessing or by enforcing invariance during model training. This raises a practical question of whether sufficient predictive signal remains once shortcut cues are suppressed. Fortunately, the availability of multi-site datasets with varying artifact distributions makes generalization approaches feasible in this setting \citep{bissoto2022artifact}.
    
    We can also choose between non-adaptive solutions (removing artifacts during preprocessing) and adaptive solutions (learning to ignore artifacts through DG techniques), which are possible given the availability of multi-site datasets with varying artifact distributions \citep{bissoto2022artifact}.
    \item \textbf{(Deployment Strategies)}
    We have now identified that shortcut features exist which we want to avoid relying on during deployment.
    One approach to deal with this problem is using counterfactual image generation to synthetically remove shortcuts from training data \citep{weng2024fast}. These methods generate modified images where markers are inpainted, allowing us to verify whether a model's predictions rely on artifacts. By training exclusively on shortcut-free images, we ensure the model learns to rely on the visual appearance of the skin lesion instead of the markers. 
    
    Alternatively, when multiple training environments with different artifact distributions are available (e.g., from different hospitals), we can employ DG methods \citep{bissoto2022artifact, yan2023epvt}. These approaches explicitly partition training data by artifact type and learn representations that suppress features correlated with environment $D$ while preserving those correlated with disease $Y$. By training to be invariant across these diverse artifact distributions, the model becomes robust to novel marking conventions encountered during deployment.
\end{enumerate}

\subsection{Case study on dementia severity prognosis}

Consider the problem of predicting \textit{Dementia Severity} in two years time. Developing models on research data, such as Alzheimer's Disease Neuroimaging Initiative\footnote{https://adni.loni.usc.edu/}, and then deploying them within real clinical settings can be challenging due to various potential data shifts. Below, we demonstrate how our causal framework can help us anticipate and act on certain data shifts that will result in a more robust predictor.

\begin{enumerate}
	\small
    \item \textbf{(Problem Conceptualization)}
    The prediction target is a probability of developing \textit{severe dementia} in two years, denoted with a binary variable $Y$. Even when an individual has been diagnosed with dementia, we may still be interested in predicting whether dementia symptoms will become \textit{severe} in two years time. Furthermore, we have access to two other important variables in our development dataset: \textit{donepezil prescription}, commonly prescribed for Alzheimer's disease symptoms~\citep{seltzer2005donepezil}, denoted with $D$, and the \textit{p-tau217 biomarker} used as reliable amyloid positivity predictor~\citep{janelidze2020cerebrospinal}, indicated with $B$.

    \item \textbf{(Causal Description)}
    At first, it might be tempting to simply model the problem as $P(Y|D,B)$, which may perform well in development. However, eliciting the data generating process in the form of the augmented causal graph will reveal potential data shifts that will degrade model's predictive performance when deployed in a clinic. Upon further examination, we may realize that donepezil prescription policy may vary between sites (e.g., conservative vs. prevention first), denoted with $S$.
    
    For instance, in a clinic with a conservative policy ($S{=}1$), donepezil may be prescribed to patients with already advanced symptoms, in which case severe dementia will occur shortly afterwards. A model trained on data from this clinic will predict high probability of $Y$ if the prescription is present ($D{=}1$). However, if the same model will be deployed within a clinic where donepezil is prescribed even when early signs of dementia are observed ($S{=}2$), it will incorrectly predict very likely severe dementia in the near future because of the prescription, even though the severe symptoms are more likely to occur only in the further future.
    
    Furthermore, we recognize that $Y$, $D$ and $B$ have the same underlying (unobserved) cause that is \textit{neurodegeneration} ($N$). The causal graph will then look as follows:
      
\begin{center}
        \includegraphics[width=6cm]{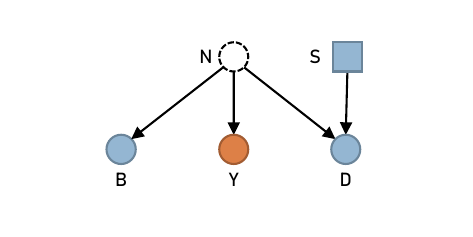}
\end{center}

    \item \textbf{(Feasibility Analysis)}
    The graph helps us identify an unstable variable $D$ due to changes in prescription policies between sites ($S$), whereas the biomarker $B$ is expected to be stable across the development and deployment environments. In this scenario, generalization is feasible by, for example, excluding the unstable variables from the model. 
    Using $D$ will improve model performance in development, but we may see the performance drop during deployment. If this drop is small, then we may be able to use $D$ anyways for deployment.  If the drop is large, then it is best to not use $D$. Knowing which case is more likely is difficult to assess, but the model could be stress tested to quantify the sensitivity to this shift. 
    Alternatively, we may consider to keep $D$ but account for the statistical difference in the prescription policy using re-weighting methods (details in Box 4), but only if we have access to deployment data before using the model in production.
    
    \item \textbf{(Deployment Strategies)}
    Given the analysis in the previous steps, our main deployment strategy is to exclude the information about donepezil prescription due to its expected instability across sites. As a result, our final model will use only the biomarker $B$ to predict dementia severity: $P(Y|B)$. If, however, deployment data are available beforehand, it is possible to model $P(Y|B,D)$ while re-adjusting $D$ to the new prescription policy.
    
\end{enumerate}

\subsection{Case study on hospital readmission}

Consider a predictive model designed to identify the risk of hospital readmission for patients with \textit{Congestive Heart Failure} ($H$). Identification of patients who are likely to be readmitted is important to allow targeted interventions that reduce mortality rates \citep{pishgar2022prediction}. A common challenge in clinical informatics is that models developed in specialized, high-acuity centers often fail when deployed in community hospitals due to differences in admission criteria. Below, we apply our causal framework to address this selection-induced dataset shift.

\begin{enumerate}
\small

    \item \textbf{(Problem Conceptualization)} 
    The prediction target is the risk of readmission with heart failure, denoted by a binary variable $Y$ (where $Y$ is therefore a proxy for the severity of $H$). We use another chronic condition, \textit{Severe Asthma} ($A$), as one of the predictor variables. To account for the shift, we include an environment variable $E$ which records the hospital's admission policy. In Environment 1 (Training), the hospital has a strict admission policy (specialized acute hospital), whereas Environment 2 (Deployment) has a more inclusive policy (community hospital).
    
    \item \textbf{(Causal Description)} 
    We must identify the structural dependencies in the data-generating process, which we visualize below. In the general population, heart failure ($H$) and asthma ($A$) are independent. However, both are causes for hospital admission ($S$). Thus, $S$ is a \textit{collider} variable \citep{HernanMonge2023Collider}.

\begin{center}
\includegraphics[width=6cm]{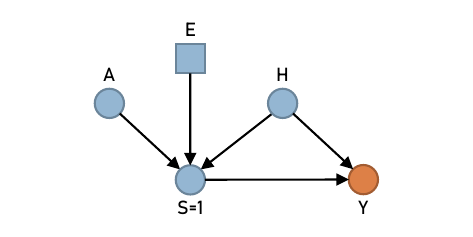}
\end{center}
    
    In our training environment, we only observe data where $S=1$. Conditioning on this collider induces a spurious correlation between $A$ and $H$ (this is known in the epidemiology literature as Berkson's bias, but is of course collider bias/selection under the null \citep{Westreich2012Berkson}). Mathematically, the observed distribution is $P(A, H | S=1, E)$. The shift occurs because the mechanism $P(S=1 | A, H)$ changes across environments $E$, altering the strength of this spurious association. This means that the predictive role of $A$ is unstable, and the model should prefer not to use this predictor.

    \item \textbf{(Feasibility Analysis)} 
    By identifying $S$ as a collider, we recognize that the relationship between $A$ and $H$ is non-structural; it is an artifact of the selection process. A standard model would learn that asthma is "protective" against heart failure readmission simply because patients without asthma must have had more severe heart failure to be admitted. Because this relationship is mediated by the selection node $S$, it is not \textit{transportable} to environments with different admission thresholds. Recovery is feasible if we can estimate the selection propensity or identify features that are d-separated \citep{HernanRobins2020WhatIf} from the selection mechanism.

    \item \textbf{(Deployment Strategies)} 
    To deal with this problem, we employ a causal re-weighting strategy. We use Inverse Probability Weighting (IPW) \citep{Chesnaye2022IPTW} to adjust the training loss, where each sample is weighted by $W = 1/P(S=1 | A, H)$. This effectively "undoes" the selection bias of the training hospital, forcing the model to ignore the spurious link between $A$ and $H$. By training on this pseudo-population, the model learns a stable relationship that remains valid even as the admission policy $E$ changes at the deployment site. This does, however, require data on those who were not admitted. An alternative would be that we use prior knowledge to tell us that $A$ and $H$ are in fact independent in the unselected population.
\end{enumerate}

\section{Conclusion}\label{sec13}
Developing reusable models in a world with data shifts requires causality. By using causal reasoning, practitioners can move from ad-hoc strategies to a principled methodology. We do not argue that causality is a panacea for all data shift problems, but rather it provides a clear line of thinking with which we can frame and propose solutions to many problems. By formalizing this process of reasoning into a four-staged framework, we build a bridge that allows machine learners and domain experts like clinicians and doctors alike to contribute to the process of building better models. Augmented causal graphs serve as a key part of this framework, providing a concise language to express diverse data shifts in a unified manner. While we acknowledge that the complexities of healthcare may extend into sociotechnical realms that evade formal mathematical descriptions, these causal descriptions provide a strong foundation for identifying when problems \textit{can} be solved. Ultimately, the value of the causal perspective is not just about proposing new tools to address data shifts, but also about knowing which tools to use in the first place.

\section*{Acknowledgments}
We acknowledge the support of the UKRI AI programme, and the Engineering and Physical Sciences Research Council, for CHAI - Causality in Healthcare AI Hub [grant number EP/Y028856/1].

\bibliographystyle{plainnat}
\bibliography{refs.bib}

\end{document}

%% file: math_commands.tex
\usepackage{amsmath,amsfonts,bm}

\def\eqref#1{equation~\ref{#1}}

\def\1{\bm{1}}

\DeclareMathAlphabet{\mathsfit}{\encodingdefault}{\sfdefault}{m}{sl}
\SetMathAlphabet{\mathsfit}{bold}{\encodingdefault}{\sfdefault}{bx}{n}

